\documentclass[letterpaper, 10 pt, conference]{ieeeconf}  

\IEEEoverridecommandlockouts                              

\usepackage{cite}
\usepackage{amsmath,amssymb,amsfonts}
\usepackage{algorithmic}
\usepackage{graphicx}
\usepackage{textcomp}
\usepackage[x11names]{xcolor}
\usepackage{fancyvrb}
\usepackage{xcolor}
\usepackage{booktabs}
\usepackage{wrapfig}
\usepackage{multirow}
\usepackage{balance}
\def\BibTeX{{\rm B\kern-.05em{\sc i\kern-.025em b}\kern-.08em
    T\kern-.1667em\lower.7ex\hbox{E}\kern-.125emX}}

\usepackage{graphicx}
\usepackage{etoolbox}
\usepackage{caption} 
\usepackage{hyperref}
\usepackage{markdown}
\usepackage{fancyvrb}

\newcommand{\insertfig}{\includegraphics[width=\linewidth]{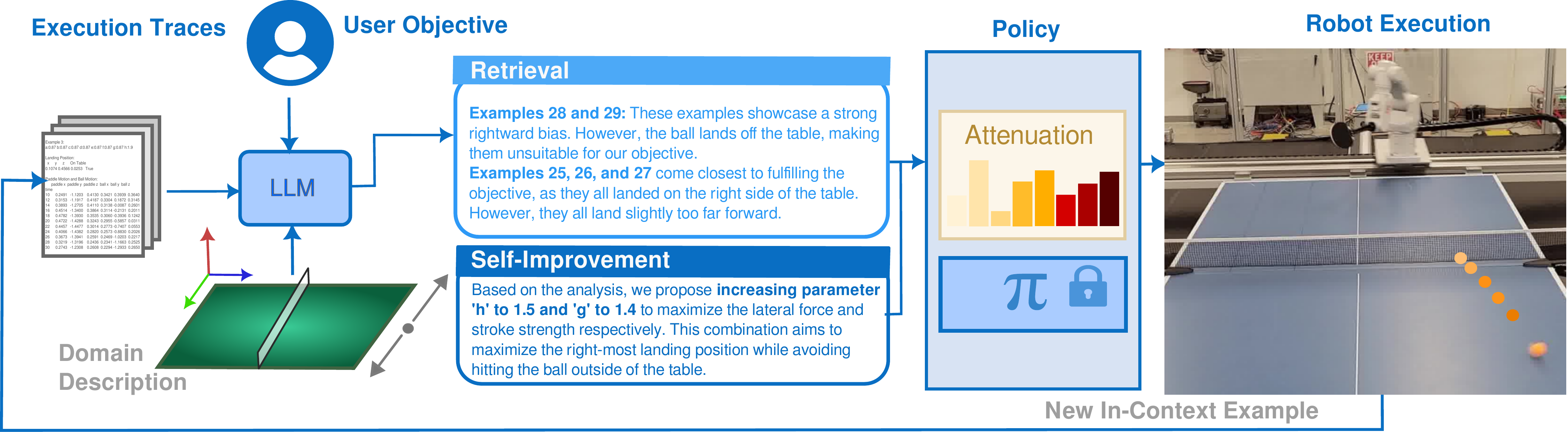}\captionof{figure}{A large language model is provided with (1) traces of robot behavior, (2) a description of the domain and (3) objectives of the human user. Using the SAS Prompt, the LLM iteratively optimizes attenuation parameters of a policy by retrieving previous examples that best fit the objectives and then synthesizing a new set of parameters in order to gradually improve robot behavior.}\label{fig:overview}}
\makeatletter
\apptocmd{\@maketitle}{\setcounter{figure}{0}\centering\insertfig}{}{}
\makeatother

\begin{document}

\title{SAS-Prompt: Large Language Models as Numerical Optimizers \\for Robot Self-Improvement}


\author{Heni Ben Amor$^{1,2}$, Laura Graesser$^{2}$, Atil Iscen$^{2}$, David D'Ambrosio$^{2}$, Saminda Abeyruwan$^{2}$, Alex Bewley$^{2}$,\\ Yifan Zhou$^{1}$, Kamalesh Kalirathinam$^{1}$, Swaroop Mishra$^{2}$, Pannag Sanketi$^{2}$ 
\thanks{$^{1}$Authors are with the School of Computing and Augmented Intelligence, Arizona State University. Corresponding author: hbenamor@asu.edu. $^{2}$Authors are  with Google DeepMind. 
}}

\maketitle

\begin{abstract}
We demonstrate the ability of large language models (LLMs) to perform iterative self-improvement of robot policies. An important insight of this paper is that LLMs have a built-in ability to perform (stochastic) numerical optimization and that this property can be leveraged for explainable robot policy search. Based on this insight, we introduce the SAS Prompt (\textbf{S}ummarize, \textbf{A}nalyze, \textbf{S}ynthesize) -- a single prompt that enables iterative learning and adaptation of robot behavior by combining the LLM's ability to retrieve, reason and optimize over previous robot traces in order to synthesize new, unseen behavior. Our approach can be regarded as an early example of a new family of explainable policy search methods that are entirely implemented within an LLM. We evaluate our approach both in simulation and on a real-robot table tennis task.  \href{https://sites.google.com/asu.edu/sas-llm/}{Project website: sites.google.com/asu.edu/sas-llm/} 

\end{abstract}

\section{Introduction}
Large language models and related foundation models~\cite{Bommasani2021FoundationModels} have a number of properties that make them particularly appealing for robotics research. One central feature is their ability to provide an intuitive natural language interface to human users. Inputs and outputs to an LLM are formatted in natural language and allow even non-expert users to interact with a robot, for example by providing new task objectives or asking for clarification. Beyond language as an interface, a variety of other, more  surprising properties have been identified such as generating chains of reasoning \cite{chain_of_thought}, complex code generation~\cite{shrivastava2023repository}, being able to solve logic problems~\cite{DBLP:conf/acl/SuzgunSSGTCCLCZ23}, and completing math puzzles~\cite{lewkowycz2022solving}, or the ability to match and complete patterns~\cite{mirchandani2023large}. As a result of these properties, LLMs have been used in robotics to generate high-level plans from human instructions~\cite{saycan}, facilitate collaboration between multiple agents~\cite{mandi2024roco}, synthesize robot policies as code~\cite{codeaspolicies2022}, or design reward functions~\cite{Liang2024LearningTL}. 


In this paper, we observe another \emph{emergent property of LLMs} that renders them particularly well-suited for robot learning and self-improvement. We demonstrate that \textbf{LLMs can effectively perform stochastic numerical optimization} out-of-the-box without the need for additional retraining. Since numerical optimization is the central backbone of many learning and adaptation algorithms, e.g. Policy Gradient methods\cite{peters2006policy}, we investigate whether LLMs can be used to perform learning through trial-and-error. Traditional methods that enable such self-improvement require a set of components such as (a) the identification of critical feature variables, (b) the design of a loss/reward function involving these features, and (c) an update rule in order to iteratively synthesize better parameters. 

\begin{figure}[t]
    \scriptsize
    \centering  
    \hrule width0.465\textwidth height0.7pt
    \vspace{0.3cm}
    \begin{Verbatim}[commandchars=\\\{\}]
\textbf{Example 5:}
a:1.1 b:1.2 c:0.7 d:1.1 e:1.1 f:1.1 g:1.1 h:1.5 

\textbf{Landing Position}:
  x       y    z      On Table
0.3207 0.7890 0.0143   True

      \textbf{paddle x  paddle y  paddle z  ball x  ball y ball z}
\textbf{time}                                                      
1      0.2478   -1.1859    0.4236  0.2415  0.0619  0.2745
2      0.2993   -1.2453    0.4059  0.2346 -0.1359  0.2134
3      0.3417   -1.2889    0.3722  0.2184 -0.3311  0.1418
4      0.3483   -1.3131    0.3347  0.2093 -0.5129  0.0451
5      0.3317   -1.3254    0.3018  0.1884 -0.6759  ...

\end{Verbatim}
\hrule width0.465\textwidth height0.7pt
\normalsize
\vspace{0.2cm}
    \caption{An example \emph{execution trace} in the robot table tennis domain. The positions of the paddle and ball at every time step are provided along with robot control parameters (a -- h).}
    \label{fig:trace}
\vspace{-0.5cm}
\end{figure}

The main contribution of this paper is the SAS Prompt -- an approach for robot learning and self-improvement that implements all three of the above steps within a single LLM prompt. \textbf{SAS Prompt enables robots to effectively understand and interpret previous robot behavior from in-context examples in order to perform policy search and synthesize new, unseen behavior.} The result is a family of algorithms in which self-improvement and numerical optimization is performed through repeated calls to an LLM with increasing context window size. An overview of the approach can be seen in Fig.~\ref{fig:overview}. The system is provided with a cache of previous robot trials in the form of execution traces -- a tabular sequence of robot states (detailed in Fig.~\ref{fig:trace}) which were observed during rollout. In turn, the LLM is tasked to generate improved robot control parameters, i.e., attenuation values, that will bring the robot behavior closer to the objectives specified by a human user. Instead of reward or fitness functions, objectives are expressed in natural language, making the process more accessible. Using its ability to process natural language and numerical data, the LLM can retrieve previous robot parameters that are in line with the human-stated objectives. In addition to this retrieval process, the LLM further analyzes the impact of the individual control parameters on task performance and, in turn, \emph{synthesizes} a new set of parameters to improve robot performance. This analysis can be regarded as a textual formulation of a gradient: understanding the effect and amplitude of involved parameters allows the LLM to make informed decisions about changes and updates. The result is an iterative process, similar in spirit to policy search and other self-improvement algorithms, which progressively generates improved control parameters. A key advantage of our approach is the explainability of individual learning steps, since the LLM provides natural language justifications for all parameter choices. We validate this behavior through extensive experiments in a robot table tennis control task in both simulation and the real-world.

\begin{figure} [t]
    \centering
    \includegraphics[width=\linewidth]{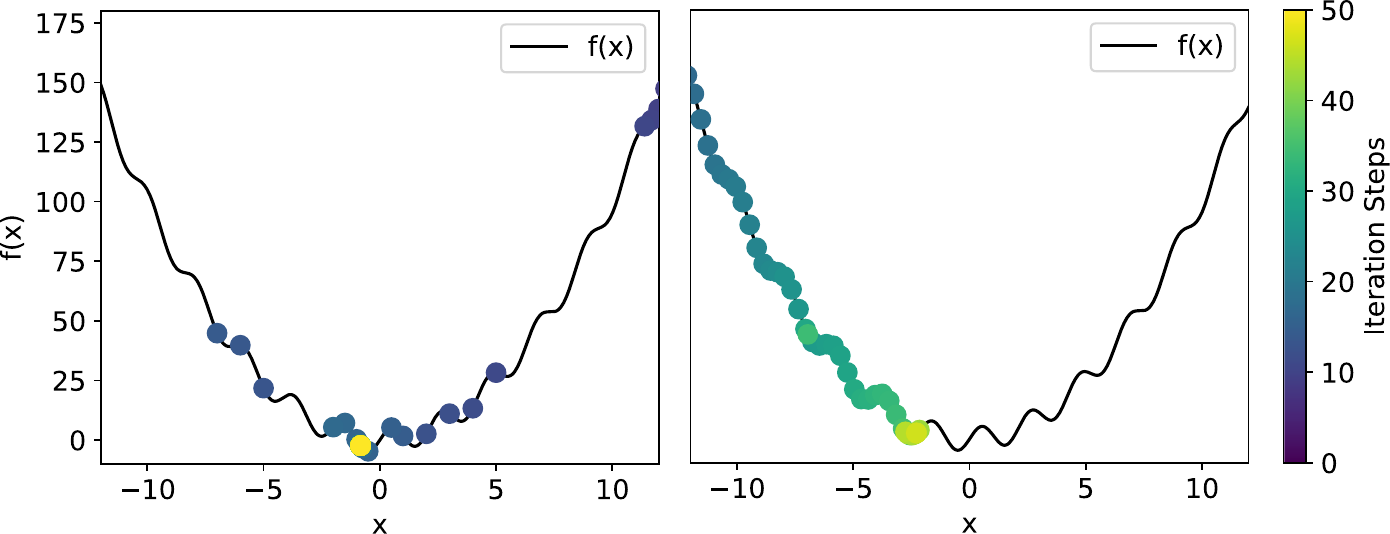}
      \scriptsize
    \centering  
    \vspace{0.02cm}
    \hrule width0.49\textwidth height0.7pt
    \vspace{0.2cm}
    \begin{Verbatim}[fontsize=\tiny, commandchars=\\\{\}]
\textbf{System Prompt:} You are an expert optimization assistant, helping me find the global
minimum of a mathematical function. I will give you the function evaluation f(x)
and the current iteration number at each step. Your goal is to  propose input 
values 'x' that efficiently lead us to the global minimum within a limited number 
of iterations (MAX_STEPS).

Here's how we'll interact:
1. I will first provide MAX_STEPS along with a few training examples of the form 
   'x, f(x)'.
2. You will provide your response in the following exact format:
    * Line 1: a new input 'x: ', aiming to minimize the function's value f(x,y).
    * Line 2: a brief explanation of why you chose that input in one sentence, 
      considering the current iteration.
3. I will then provide the function's value f(x) at your suggested point, and the 
   current iteration.
4. We will repeat steps 2-3 until we reach the maximum number of iterations.

Remember:
* **Assume no prior knowledge about the function's specific form.**
* **Balance Exploitation and Exploration:**  Early on, explore the input space 
  broadly. As iterations increase, focus more on promising regions based on 
  observed f(x) values.
* **Be adaptable:**  Your approach might need to change based on the function's 
  behavior and the remaining iterations. If you suspect a local minimum or slow 
  progress for too long, try more exploratory values and then exploit any 
  promising findings based on your understanding of the function.

\textbf{----- LLM Numerical Optimization Process: }

\textbf{Step:} 0, x: -4.5
Explanation: We observe a decreasing trend in f(x) as x increases. Let's continue
moving in the positive x direction, taking a slightly larger step to accelerate
the search.
f(x): 18.189 % feedback to LLM from user

\textbf{Step}: 1, x: -2.5, 
Explanation: The function value decreased significantly. It seems like moving with 
bigger steps is a good strategy for now - hence another relatively big jump.
f(x): 11.045 % feedback to LLM from user
…
\textbf{Step}: 13, x: -0.715 
Explanation: The function values at x = -0.719 and x = -0.711 are practically the 
same. There is a high chance that the minimum lies between these points or very 
close to them. Let's investigate the midpoint for a potentially better estimate.
f(x): -4.439  % feedback to LLM from user
\end{Verbatim}
\hrule width0.49\textwidth height0.7pt
\normalsize
\vspace{0.2cm}
  \caption{\textbf{Numerical Optimization inside an LLM}: the top figures depict two runs of an optimization process leveraging an LLM. The bottom text shows how the LLM is iteratively queried to minimize function $f(x)$. No gradients are provided. \label{fig:optimization}}
\vspace{-0.7cm}
\end{figure}

\section{Related Work}
Neural network approaches for incorporating language understanding and action generation have a storied tradition in robotics~\cite{wermter2003learning, NIPS2003_3baa271b}. Yet, a recent resurgence in interest can be attributed to the introduction of efficient embedding techniques~\cite{Mikolov2013EfficientEO} and large language models~\cite{team2023gemini, OpenAI_GPT4_2023}. For example, the works~\cite{stepputtis2020language, Lynch-RSS-21} discussed new methodologies for imitation learning of \emph{vision-language} models that enable robots to interpret human language. Going beyond interpretation, the SayCan~\cite{saycan} framework enables robots to leverage the reasoning capabilities of LLMs in order to decompose a task and generate executable plans. A variety of works have followed-up on this methodology~\cite{huanginner, song2023llm, ding2023task} expanding on these planning and reasoning capabilities. To incorporate learning, the work in~\cite{Liang2024LearningTL} uses an LLM to translate human objectives into a cost function. In turn, a model-predictive planner finds optimal actions that minimize the generated cost function. In a similar vein, the Eureka system~\cite{ma2024eureka} uses evolutionary optimization over reward code in conjunction with reinforcement learning to acquire complex motor skills. Several works have explored the concept of \emph{in-context learning}~\cite{fu2024icrt, Palo-RSS-24, liang2024learninglearnfasterhuman}. In these approaches, human demonstrations, corrections or feedback are used by the LLM to refine or generalize a robot skill. Outside of robotics, FunSearch~\cite{funsearch} exploits LLM  \emph{hallucinations} in order to discover new mathematical insights. However, all of these approaches assume either (1) an optimization loop outside of the LLM or (2) human expert demonstrations that already solve the task. In our paper, we are particularly interested in the question of whether language models are able to adapt and self-improve a robot policy without the need for an external optimization loop or repeated coaching from a human, i.e., only robot execution and data logging are performed outside of the LLM. A key insight is that such models have emergent capabilities for performing numerical optimization. Similar emergent properties have previously been reported for prompt optimization~\cite{2309.03409v3} and hyper-parameter optimization~\cite{zhang2023using}. For example, the work in~\cite{2309.03409v3} demonstrated that LLMs can iteratively be queried to perform prompt optimization. The same paper also reports results on the traveling salesman task -- a widely known NP-hard discrete optimization task. By contrast, in this paper, we focus in particular on robot policy parameters and investigate the suitability of LLMs for the iterative optimization of such numerical values.


\section{Robot Table Tennis}
Without loss of generality, we discuss our methodology within the context of robot table tennis~\cite{mulling2013learning}. For real-world experiments we use a 6 DoF arm mounted on top of two linear gantries, enabling motion in the 2d plane. A detailed description of the setup (8DoF in total) can be found in~\cite{D'Ambrosio-RSS-23, abeyruwan2023sim2real}. In~\cite{d2024achieving} we presented a robot that plays competitively against a human opponent. By contrast, in this paper, we are interested in the robot's ability to translate human targeting instructions into control parameters. Rather than immediately controlling the robot actions, the LLM generates attenuation parameters for a lower-level controller (LLC). This LLC is able to hit the ball but may not be able to (a) reliably land it, (b) target a landing location, or (c) interpret human language instructions. We aim to modulate this simple LLC to enable language-conditioned robot control through retrieval and self-improvement of attenuation parameters. Attenuation parameters are similar in spirit to a residual layer, in which each of the outputs of the LLC are multiplied by linear scaling factor. The robot actions $\mathbf{x} \in \mathbb{R}^8$ generated by the LLC correspond to the velocities of the eight robot actuators. These robot actions are then multiplied with an attenuation vector $\mathbf{\theta}\in \mathbb{R}^8$ of the same length to yield the final robot action. We will discuss in this paper how this attenuation vector is generated through interaction with an LLM. 

\section{Methodology}


\begin{figure}[t]
    \centering
    \includegraphics[width=\linewidth]{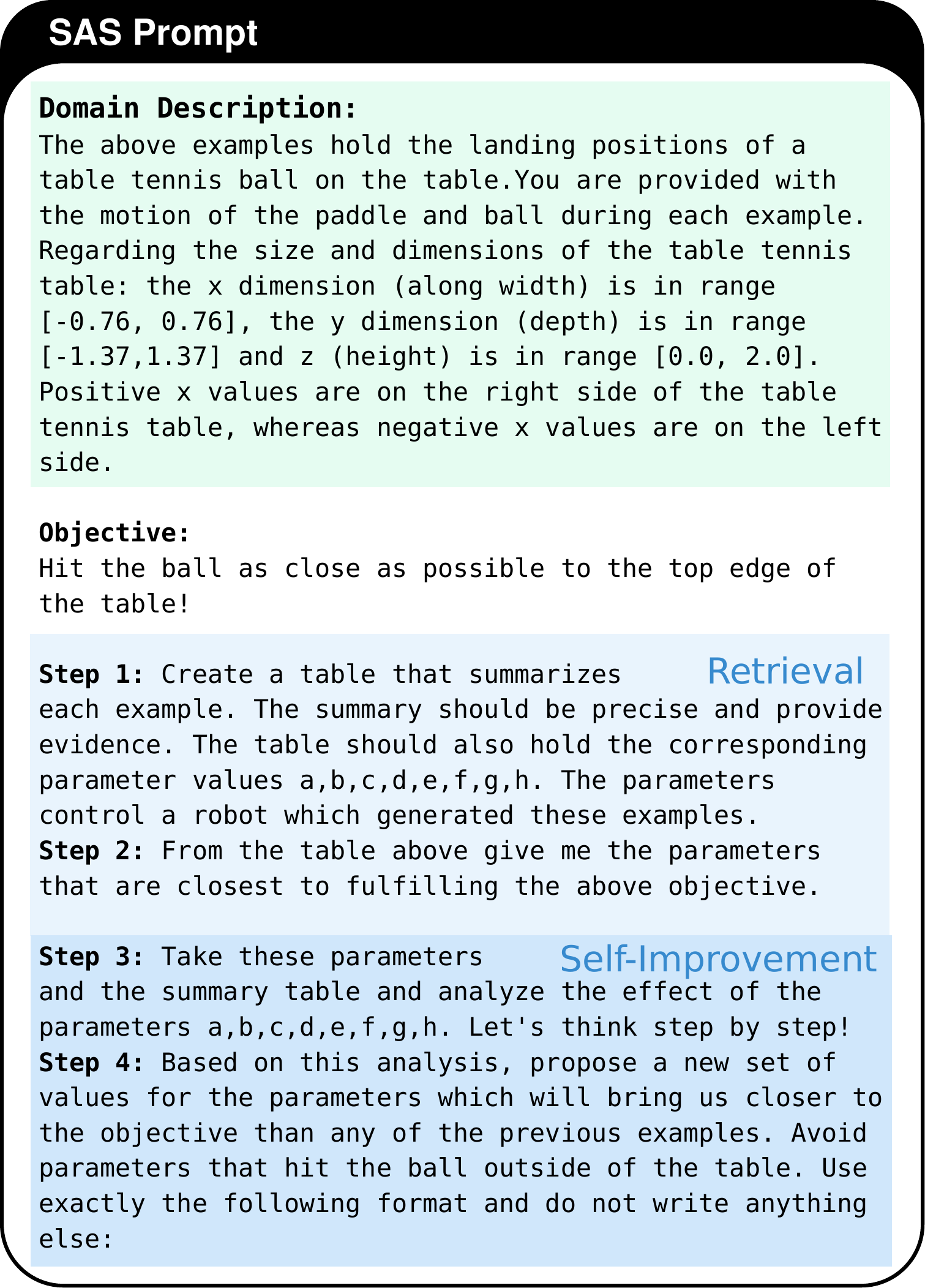}
    \caption{\textbf{SAS Prompt}: the prompt provides the LLM with information about the domain, the user objectives and a step-by-step instruction on how to summarize and analyze the in-context examples. In turn, the final step is to synthesize a new set of parameters. Step 1 and 2 aim at retrieving previous examples that are best aligned with human objective. Step 3 and 4 aim at further optimizing these values to improve robot performance with regards to the same objective.}
    \label{fig:sasprompt}
\vspace{-0.15in}
\end{figure}

In this section we motivate our approach and describe the SAS Prompt. A cornerstone of our methodology is our observation that LLMs are inherently capable of numerical optimization. Fig.~\ref{fig:optimization} shows a general example of how the LLM can operate as a numerical optimizer to determine the global minimum of non-convex functions. To this end, the LLM is repeatedly asked to provide the next candidate for input parameter $x$ that would lead to the highest reduction in function value. Neither the function $f$ nor its derivatives are known to the LLM. Instead, only the corresponding function values $f(x)$ are provided whenever it proposes a new candidate value for $x$. The result is a gradient-free optimization process that unfolds as a result of the repeated interaction with the LLM. Due to its natural language capabilities, the LLM can also be instructed to \emph{explain} the parameter choices made at each step, thereby increasing the transparency of the underlying process. 

\subsection{The \textbf{S}ummarize, \textbf{A}nalyze, \textbf{S}ynthesize Prompt}

Building upon the above insight, we introduce the SAS Prompt -- a prompting technique combining retrieval and numerical optimization that enables learning and adaption from previous robot experiences. The SAS prompt realizes three objectives that allow the LLM to identify a set of robot control parameters:

\begin{itemize}
\item {\large\textbf{S}}ummarize: all in-context examples are summarized and important features are extracted. This step forces the language model to process each example.
\item {\large\textbf{A}}nalyze: the impact of control variables on robot behavior is analyzed in order to identify correlations and dependencies. These insights inform the later parameter synthesis process.
\item {\large\textbf{S}}ynthesize: a set of control parameters is generated. This can be done through (a) pure retrieval, i.e., retrieve the best fitting parameter sets in the cache or, (b) by synthesizing a completely new set of parameters.
\end{itemize}
After querying the LLM using the SAS prompt, the proposed set of control parameters is executed on the robot yielding a new execution trace. The new trace of robot behavior is appended to the current LLM in-context cache in order to allow for incremental improvement and learning from prior experiences. Fig.~\ref{fig:sasprompt} depicts the SAS Prompt in the context of the robot table tennis domain. The prompt can logically be divided into four main components: 


\noindent
\textbf{A. Domain Description and User Objective:} The first component of the SAS prompt is a description of the domain, along with the underlying task and environment. This component provides the necessary information needed to interpret the in-context examples and perform spatial reasoning. An essential element of this description is a definition of the coordinate system inherent to the task, e.g., the orientation and meaning of reported coordinate dimensions along with their ranges. In the example in Fig.~\ref{fig:sasprompt}, we specify the overall task (table tennis) as well as the table dimensions. The second component of the SAS prompt is the user objective. Here the instructions of the human are provided, e.g., ``Hit to the top edge of the table". 

\noindent
\textbf{B. Retrieval}: In the retrieval section of the SAS prompt, the LLM is first tasked with \emph{summarizing} all in-context examples within a table. This step directs the LLM to inspect each example and extract critical features and insights before reaching a conclusion. \begin{wrapfigure}{r}{0.25\textwidth}
    \centering
    \includegraphics[width=\linewidth]{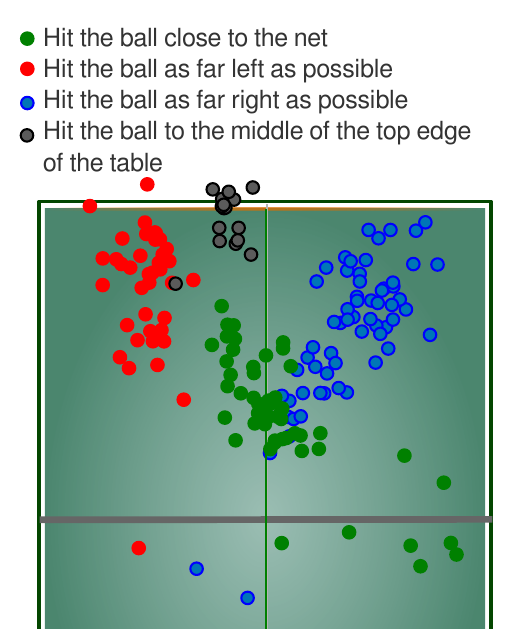}
    \caption{\textbf{Retrieval} results: the retrieval part of the SAS prompt is used to retrieve robot control parameters which best fit the human instructions (printed above). Depicted coordinates are ball landing locations in a real-robot experiment.}
    \label{fig:retrieval}
\end{wrapfigure}The specific categories of the table may be provided with the prompt or can be omitted. In practice, we observe that the LLM will automatically choose fields that are best suited to answering the human query. The second step of the SAS prompt directs the LLM to identify in the table the examples / control parameters that are closest to fulfilling the given objective. This is a critical step since it requires the LLM to reason about the summarized table and identify one or multiple possible candidates. In traditional methods for self-improvement, this step is the result of applying the fitness or reward function to the collected samples and then selecting the top set of performing candidates for further adjustment.



\noindent
\textbf{C. Self-Improvement:} The final component of the SAS prompt is self-improvement. The LLM is tasked with analyzing the effect of the variables on the in-context examples. The underlying goal is to identify the impact of each variable on the execution trace. For each of the robot control variables, the LLM identifies possible correlations and dependencies. Understanding these relationships are critical to the subsequent adjustment step. At its core, the analysis can be seen as a hypothesis that is refined with every iteration. Based on this hypothesis, the LLM is then tasked with generating a new set of parameters that will bring the robot closer to achieving the task objectives than any one of the previous examples. This step mimics the prompt in the numerical optimization example introduced in Fig.~\ref{fig:optimization}. Every new parameter setting proposed by the LLM can be regarded as a sample in a gradient-free optimization process. Through the synthesis of gradually better parameter values, the LLM is implicitly executing an optimization process in which the cost function is provided in natural language. Note also, that the LLM can explicitly be asked to justify these choices in a text. 

\begin{figure} [t]
    \centering
    \includegraphics[width=0.9\linewidth]{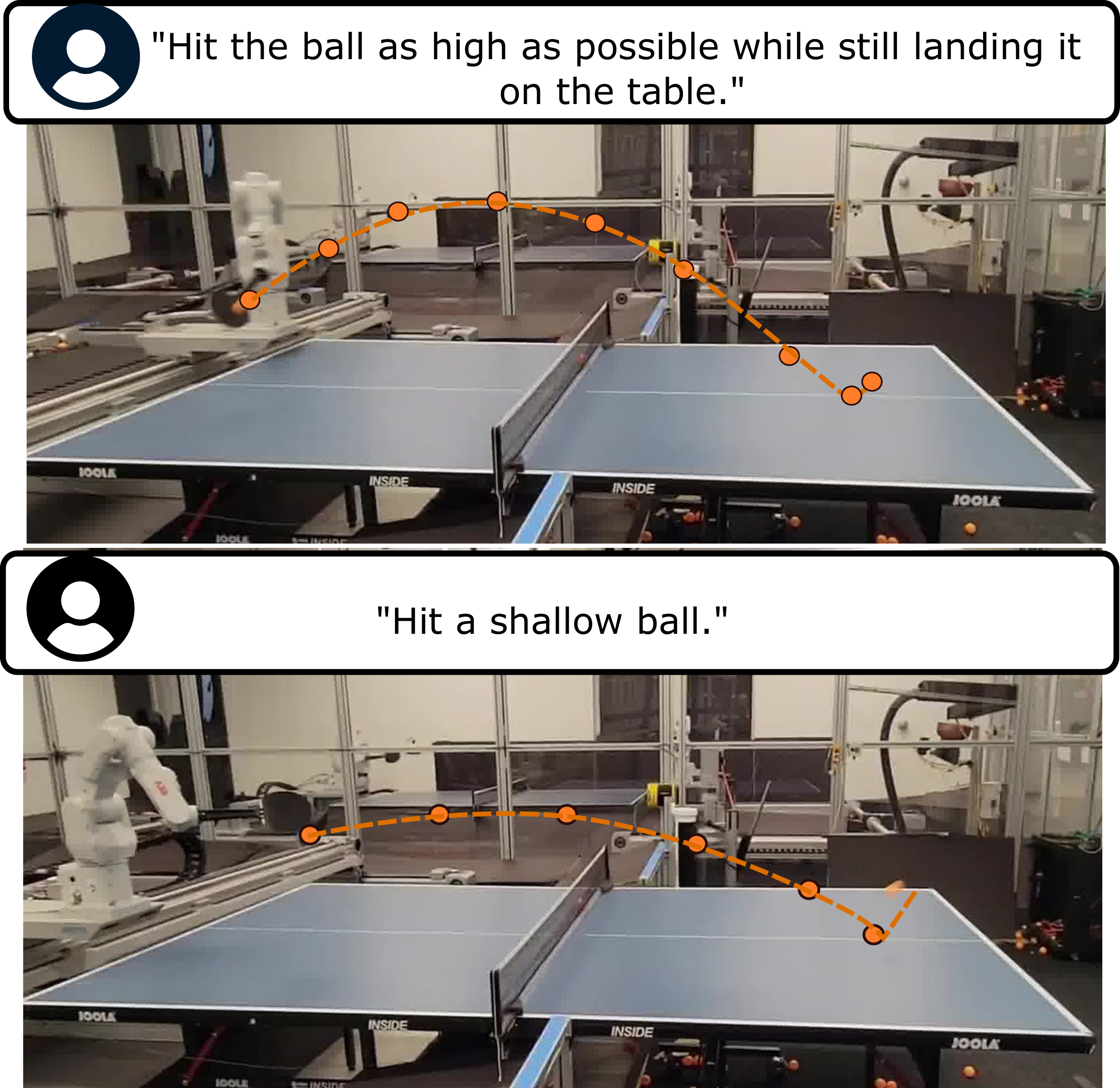}
    \caption{Visualization of the ball trajectory in a retrieval task. }
    \label{fig:highlow}
\vspace{-0.15in}
\end{figure}


\section{Evaluation and Results}


\subsection{Numerical Optimization inside the LLM}
One of the key tenets of this paper is that LLMs are capable of numerical optimization and that this property can be exploited for self-improvement in robotics. In order to provide evidence, we conduct a set of standard experiments for numerical optimization and compare the performance of Gemini 1.5 Pro with (a) gradient descent, (b) Adam \cite{KingBa15}, (c) Nelder-Mead \cite{nelder1965simplex}, and (d) random search~\cite{rastrigin1963convergence}. The experiment does not involve the full SAS prompt, but rather the minimal methodology described in the early example in Fig.~\ref{fig:optimization}. We evaluate on widely used optimization functions (minimization), namely Rastrigin's~\cite{rastrigin1963convergence} and Ackley's function~\cite{ackley2012connectionist}, in both 2D and 8D. To avoid any potential for guessing the answer (e.g., the LLM generating (0, 0) due to this being frequently used as location for optima in benchmarks) we added a constant shift to both functions. For every algorithm 50 experiments are performed, each with 100 update steps. Table~\ref{tab:numerical} summarizes the results, as well as the initial values $f(x)$ at start of optimization. We observe that the LLM favorably compares to other methods, including Adam and Nelder-Mead. Surprisingly, LLM performance is particularly strong with more dimensions. In general, our results confirm the hypothesis that LLMs are capable of optimization. It is also important to note that both gradient descent and Adam have access to gradients or perform additional steps to approximate a gradient whereas the above LLM-based results are completely gradient-free.    

\begin{table} 
\footnotesize
\begin{tabular}{l l l l l l} 
 \toprule
 Alg. & 2D Ackley & 2D Rastr. & 8D Ackley & 8D Rastr. \\ [0.5ex] 
 \midrule
Init $f(x)$ & 11.26$\pm$3.36    & 44.07$\pm$22.78 & 17.20$\pm$1.42 & 598.80$\pm$180.57  \\
GD          & 9.49$\pm$ 3.52    & 24.40$\pm$20.90 & 15.59$\pm$1.39 & 505.29$\pm$182.62\\ 
Adam        & 6.22$\pm$5.31     & 23.14$\pm$19.59 & 15.10$\pm$2.14 & \textbf{268.93}$\pm$120.43\\ 
Neld.-M.   & 8.14$\pm$4.92     & 23.26$\pm$19.83 & 15.54$\pm$1.67 & 426.42$\pm$187.84\\
Rand.-S.    & 11.15$\pm$3.45    & 43.35$\pm$25.20 & 17.15$\pm$1.44 & 589.04$\pm$182.92\\
LLM        & \textbf{4.38}$\pm$3.15 & \textbf{9.02}$\pm$8.51 & \textbf{13.71}$\pm$3.30 & 340.87$\pm$156.47 \\ 

 \bottomrule
\end{tabular}
\caption{Performance of different numerical optimization methods compared to LLM-based optimization.\label{tab:numerical}}
\normalsize
\vspace{-0.5cm}
\end{table}

\begin{figure} [b]
    \centering
    \includegraphics[width=\linewidth]{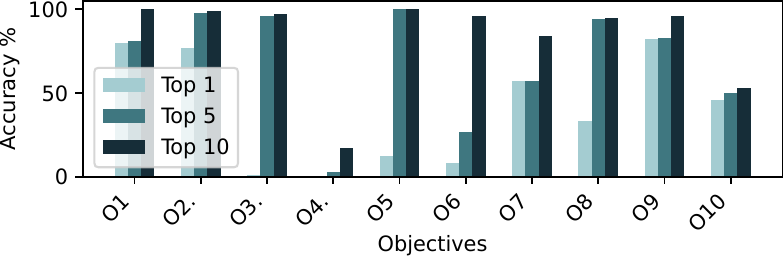}
    \caption{Results of retrieval experiments with 10 predefined queries run 100 times through the SAS prompt. For each query we identify the ground truth and evaluate LLM retrieval performance.}
    \label{fig:evalretrieval}
\end{figure}

\begin{figure*} [t]
    \centering
    \includegraphics[width=\linewidth]{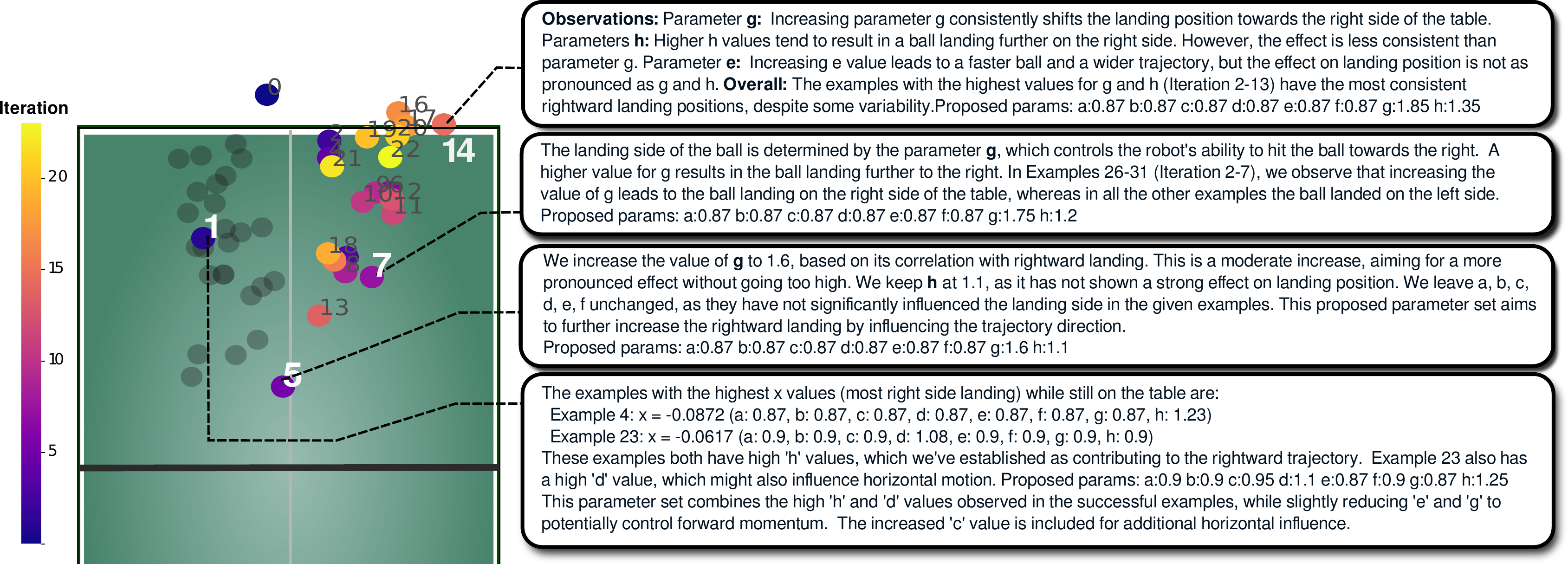}
    \caption{\textbf{Self-Improvement via SAS}: transparent points show in-context examples at start; all sampled from a narrow area on the left side (transparent points). The task is to \textbf{Hit the ball as far right as possible}. Text boxes show responses at iterations.}
    \label{fig:selfimprove-explain}
\vspace{-0.15in}
\end{figure*}

\subsection{SAS Retrieval Experiments: Real Robot}
In this experiment we evaluate the ability of SAS prompt to retrieve the correct samples given an instruction from a human user. More specifically, we provide a set of potential user objectives and evaluate the ability of the SAS prompt in identifying the best-fitting in-context example. Hence, in this experiment, we do not use the self-improvement component of the SAS prompt and focus on the retrieval aspects of the task. The experiment is conducted by providing a set of instructions to the LLMs and recording the overall ball and robot motion. The LLM has access to 100 in-context examples. Fig.~\ref{fig:retrieval} depicts the landing locations of the ball. Note that the non-deterministic nature of the domain introduces noise into these results. Despite the relative lack in precision, we notice that a general tendency to correctly translate the human instructions into real-world actions. Especially in the case of hitting to the top edge of the table, we observe a faithful implementation of the commanded objective. In addition to above results, Fig.~\ref{fig:highlow} shows another set of instructions in which the robot is instructed to modulate the height of the hit. A particular challenge in this task, is that the LLM is required to identify the peak height of each in-context example. This requires a careful analysis the provided ball trajectories in order to determine the peak height and include it as part of the summary table in step 1 of the SAS prompt. To empirically evaluate the retrieval capabilities of our approach, we further performed a systematic evaluation in which 10 different instructions are provided to the LLM in 100 independent experiments. For each of these queries we identify the ground truth by programmatically inspecting the full set of available in-context examples. More specifically, we run the following objectives through SAS: (\textbf{O1}) ``\emph{Play as far right as possible}", (\textbf{O2}) ``\emph{Aim the ball at the leftmost edge of the table}'', (\textbf{O3}) ``\emph{Play the ball close to the net}", (\textbf{O4}) ``\emph{Land the ball in the middle of the opponent's side of the table}" (\textbf{O5}) ``\emph{Target the back-left corner of the opponent's court}", (\textbf{O6}) ``\emph{Aim the ball as close as possible to the back edge of the opponent's court}", (\textbf{O7}) ``\emph{Make sure to land the ball as close as possible to coordinate [0.2, 0.8, 0.0]}", (\textbf{O8}) ``\emph{Hit the ball as close as possible to coordinate [-0.2, 0.8, 0.0]}", (\textbf{O9}) ``\emph{Play the balls so it achieves the maximum peak height while still landing it on the table}", (\textbf{O10}) ``\emph{Play as shallow a ball as possible}". Fig.~\ref{fig:evalretrieval} shows the results of this analysis. On average the LLM achieves $39.4\%$ (Top-1), $68.89\%$ (Top-5), and $83.70\%$ (Top-10). In general, we can observe a tendency of the LLM to generate reasonable responses, even if the top-1 response is found in about $~40\%$ of cases. However, strong outliers are objectives O3 and O4 which produce a Top-1 score of less than $10\%$. This might be attributed to the ambiguity of the instruction, since the middle of the table is not a single point but rather an area. 

\subsection{SAS Self-Improvement Experiments: Real Robot}
\begin{figure} [t]
    \centering
    \includegraphics[width=\linewidth]{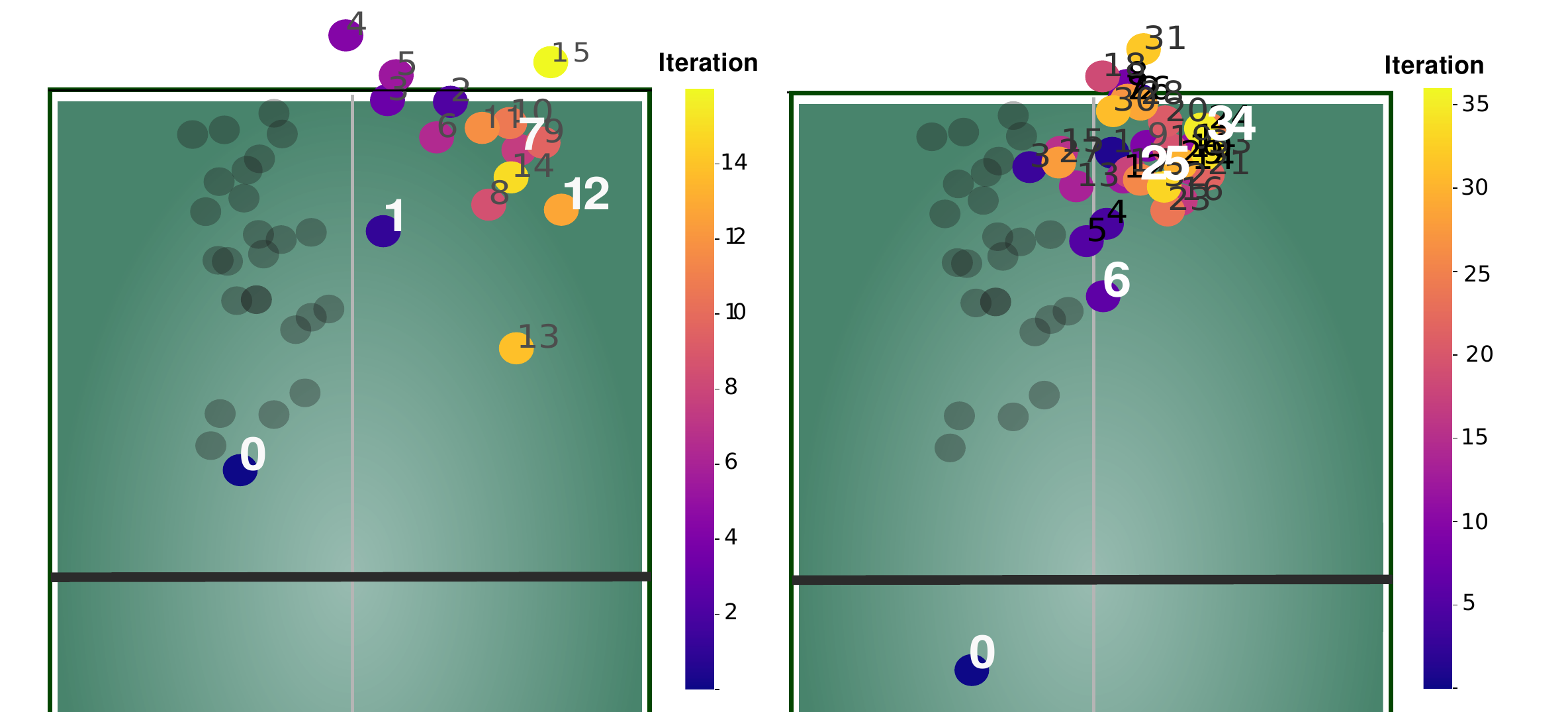}
    \caption{\textbf{Self-Improvement Results:} visualization of the learning process during the self-improvement task on a real-robot. The objective is to learn to hit to the far right side.\label{fig:addrobotexp}}
\vspace{-0.15in}
\end{figure}

In this experiment, we demonstrate the self-improvement capabilities within the robot table tennis domain. We initialize the in-context examples with 24 examples which cover a narrow stretch of the left side of the table. The user objective is specified as ``\textbf{Hit the ball as far right as possible!}". Since the in-context examples do not include any sample landing on the right side, the LLM needs to reason about the parameters and their influence on the landing site of the ball. At each iteration of the self-improvement process, an automated ball launcher system propels the ball in direction of the forehand of the robot. The robot executes its actions using the latest set of control parameters (attenuation values). In turn, the newly recorded execution trace is appended to the in-context examples and the SAS prompt is sent to the LLM in order to synthesize the next set of proposed parameters. The process is repeated in order to gradually improve the robot behavior. Fig.~\ref{fig:selfimprove-explain} shows the result of performing this process on the physical robot system. The transparent points in the figure indicate the in-context training samples available to the LLM at the beginning of the self-improvement process. We notice that within a small number of iterations the right side of the table is reached. By iteration 14, the robot is able to reliably hit to the (far) right side of the table. The figure also includes the response of the LLM at different steps of the learning process. We notice that the LLM attempts to identify correlations between individual parameters and the rightward landing objective. By iteration 14, it observes that ``\textbf{Increasing parameter g consistently shifts the landing position towards the right side of the table. Parameters h: Higher h values tend to result in a ball landing further on the right side. However, the effect is less consistent than parameter g."}. The analysis can be regarded as a textual gradient -- it relates changes in parameter values to changes in task objectives and outcomes. This textual formulation of a gradient, in turn, informs the synthesis/update of new parameter values as can be seen in the provided justification at iteration 1, e.g., ``\textbf{This parameter set combines the high 'h' and 'd' values observed in the successful examples, while slightly reducing 'e' and 'g' to potentially control forward momentum. The increased 'c' value is included for additional horizontal influence.}". Please note that the LLM is articulating the real-world implications of parameter changes, i.e., forward momentum and horizontal influence, thereby attempting to uncover their inherent meaning. Additional self-improvement experiments on the real robot are shown in Fig.~\ref{fig:addrobotexp}.



\subsection{SAS Self-Improvement: Simulation Results}\label{AA}
We also conducted a larger set of experiments using a publicly available robot table tennis simulation~\cite{fancy_gym} in Mujoco. We provided three different self-improvement objectives, namely (\textbf{S1}) \emph{``Hit the ball to the far right!"}, (\textbf{S2}) \emph{``Hit the ball to the top edge!"} and (\textbf{S3}) \emph{``Hit the ball to the left corner!"} and ran 20 experiments for each objective. In each experiment, 30 iterations of self-improvement are conducted. In S2 and S3, we provided only in-context examples from the lower half of the table. Table~\ref{tab:simulationres} holds the results of this experiment. Depicted are the mean distance of the in-context examples from the goal at the start of the experiment, as well as the distance of the landing positions after self-improvement. Additionally, it also shows the median position from the final landing positions after self-improvement. We observe that in S1 the median X position is above $0.5$m, showing a clear trend towards the right side of the table. In S2 and S3 the Y position is above $1.2$m indicating a shift towards the top edge of the table (the edge of the table being at Y = $1.37$m). Only in S3 is the X position shifted to the left of the table after training, i.e., negative X dimension. These results show that all three objectives were successfully realized with substantial changes in ball landing positions.


\begin{table} 
\footnotesize
\centering
\begin{tabular}{l | c c | c c | c c c} 
 \toprule
 \multirow{2}{*}{Objective} & \multicolumn{2}{c|}{Init. Dist.} & \multicolumn{2}{c|}{Final Dist.} & \multicolumn{2}{c}{Median Land Pos} \\ 
  & Mean & Std. & Mean & Std. & X & Y \\ [0.5ex] 
 \midrule
 \textbf{S1}: Right & 0.677 & 0.156 & 0.231 & 0.070 & \colorbox{CadetBlue2}{0.511} & 0.790 \\ 
 \textbf{S2}: Top & 1.040 & 0.086 & 0.270 & 0.295 & -0.213 & \colorbox{CadetBlue2}{1.233} \\
 \textbf{S3}: Left Cor. & 1.277 & 0.147 & 0.342 & 0.201 & \colorbox{CadetBlue2}{-0.529} & \colorbox{CadetBlue2}{1.227}	\\ 
 \bottomrule
\end{tabular}
\caption{Distance to the goal before and after training and median position are shown. All numbers are shown in meters. \label{tab:simulationres}}
\normalsize
\vspace{-0.7cm}
\end{table}

\section{Conclusion}
In this paper, we introduced the SAS prompt -- a prompting technique for robot self-improvement and  adaptation through explainable updates to policy parameters. Our approach provides a pathway towards new types of robot learning algorithms that are fully implemented within foundation models, e.g., all elements of the learning loop occur within the LLM. For future work, we hope to investigate scaling up the approach to higher dimensionality tasks and environments.


\bibliographystyle{unsrt}
\balance
\bibliography{references}

\end{document}